\documentclass[11pt,a4paper]{article}
\usepackage[utf8]{inputenc}
\usepackage[T1]{fontenc}
\usepackage{amsmath,amssymb}
\usepackage{graphicx}
\usepackage{booktabs}
\usepackage{array}
\usepackage[margin=2.2cm]{geometry}
\usepackage[hidelinks]{hyperref}
\hypersetup{pdftitle={Physics-Shaped, Not Physical: Auditing Physics-Guided Underwater Image Restoration},
  pdfauthor={Mochamad Rizal Fauzan, Irgi Surya},
  pdfkeywords={underwater image restoration; physics-guided deep learning; physical validation; data efficiency}}
\usepackage{microtype}
\usepackage{enumitem}
\usepackage{xcolor}
\usepackage[font=small,labelfont=bf,labelsep=period,justification=justified,singlelinecheck=false]{caption}

\newcommand{\NImgAUIED}{3000}

\newcommand{\NImgLSUI}{4279}

\newcommand{\MeanUIEBFullPSNR}{23.17}

\newcommand{\MeanUIEBNoPhysPSNR}{22.81}

\newcommand{\MeanLSUINoTaskPSNR}{21.44}

\newcommand{\MeanLSUIUnetPSNR}{21.60}

\newcommand{\MeanAUIEDNoPhysPSNR}{19.59}

\newcommand{\MeanAUIEDUnetPSNR}{19.66}

\newcommand{\DiffUIEBFullVsNoPhysPSNR}{+0.37}
\newcommand{\LoUIEBFullVsNoPhysPSNR}{+0.24}
\newcommand{\HiUIEBFullVsNoPhysPSNR}{+0.50}
\newcommand{\PsUIEBFullVsNoPhysPSNR}{0.046}

\newcommand{\DiffLSUIFullVsNoPhysPSNR}{\ensuremath{-}0.06}
\newcommand{\LoLSUIFullVsNoPhysPSNR}{\ensuremath{-}0.28}
\newcommand{\HiLSUIFullVsNoPhysPSNR}{+0.17}

\newcommand{\DiffLSUIFullVsUnetPSNR}{\ensuremath{-}0.06}
\newcommand{\LoLSUIFullVsUnetPSNR}{\ensuremath{-}0.25}
\newcommand{\HiLSUIFullVsUnetPSNR}{+0.12}

\newcommand{\DiffLSUIFullVsNoTaskPSNR}{+0.09}

\newcommand{\DiffLSUIsubFullVsNoTaskPSNR}{+0.10}
\newcommand{\LoLSUIsubFullVsNoTaskPSNR}{\ensuremath{-}0.05}
\newcommand{\HiLSUIsubFullVsNoTaskPSNR}{+0.25}

\newcommand{\DiffAUIEDFullVsNoPhysPSNR}{+0.06}
\newcommand{\LoAUIEDFullVsNoPhysPSNR}{\ensuremath{-}0.51}
\newcommand{\HiAUIEDFullVsNoPhysPSNR}{+0.62}

\newcommand{\DiffAUIEDFullVsUnetPSNR}{\ensuremath{-}0.01}
\newcommand{\LoAUIEDFullVsUnetPSNR}{\ensuremath{-}0.72}
\newcommand{\HiAUIEDFullVsUnetPSNR}{+0.70}

\newcommand{\DiffAUIEDFullVsNoTaskPSNR}{0.00}

\newcommand{\ParamsFull}{9.82}
\newcommand{\GflopsFull}{32.2}

\newcommand{\ParamsUnet}{7.57}
\newcommand{\GflopsUnet}{9.6}
\newcommand{\EpochsFull}{76}
\newcommand{\EpochsNoPhys}{67}
\newcommand{\EpochsNoTask}{67}
\newcommand{\EpochsUnet}{86}
\newcommand{\STNimg}{411}
\newcommand{\STRhoT}{\ensuremath{-}0.07}
\newcommand{\STRhoNull}{\ensuremath{-}0.10}
\newcommand{\STDeltaT}{+0.02}

\newcommand{\STRhoB}{\ensuremath{-}0.05}
\newcommand{\STRsq}{0.4}
\newcommand{\STFracNeg}{55}
\newcommand{\STFracNull}{60}
\newcommand{\STFracDCP}{62}
\newcommand{\STFracBpos}{40}
\newcommand{\STWilcoxP}{0.002}
\newcommand{\CastInput}{0.70}
\newcommand{\CastFull}{0.49}
\newcommand{\CastNoPhys}{0.47}
\newcommand{\CastUnet}{0.49}

\newcommand{\DFDiffTenUIEB}{+0.65}
\newcommand{\DFLoTenUIEB}{+0.06}
\newcommand{\DFHiTenUIEB}{+1.24}

\newcommand{\DFDiffTenLSUI}{+0.17}
\newcommand{\DFLoTenLSUI}{\ensuremath{-}0.38}
\newcommand{\DFHiTenLSUI}{+0.73}

\newcommand{\DFDiffTenAUIED}{+1.48}
\newcommand{\DFLoTenAUIED}{\ensuremath{-}0.08}
\newcommand{\DFHiTenAUIED}{+3.03}

\newcommand{\DFDiffTwentyFiveUIEB}{+0.57}
\newcommand{\DFLoTwentyFiveUIEB}{\ensuremath{-}0.35}
\newcommand{\DFHiTwentyFiveUIEB}{+1.48}

\newcommand{\DFDiffTwentyFiveLSUI}{+0.15}
\newcommand{\DFLoTwentyFiveLSUI}{\ensuremath{-}0.28}
\newcommand{\DFHiTwentyFiveLSUI}{+0.59}

\newcommand{\DFDiffTwentyFiveAUIED}{+0.41}
\newcommand{\DFLoTwentyFiveAUIED}{\ensuremath{-}0.34}
\newcommand{\DFHiTwentyFiveAUIED}{+1.16}

\newcommand{\DFDiffFiftyUIEB}{+0.19}
\newcommand{\DFLoFiftyUIEB}{\ensuremath{-}0.66}
\newcommand{\DFHiFiftyUIEB}{+1.05}

\newcommand{\DFDiffFiftyLSUI}{\ensuremath{-}0.07}
\newcommand{\DFLoFiftyLSUI}{\ensuremath{-}0.67}
\newcommand{\DFHiFiftyLSUI}{+0.53}

\newcommand{\DFDiffFiftyAUIED}{+0.21}
\newcommand{\DFLoFiftyAUIED}{\ensuremath{-}0.43}
\newcommand{\DFHiFiftyAUIED}{+0.85}

\newcommand{\DFDiffHundredUIEB}{+0.37}
\newcommand{\DFLoHundredUIEB}{+0.24}
\newcommand{\DFHiHundredUIEB}{+0.50}
\newcommand{\DFPHundredUIEB}{0.017}

\newcommand{\DFDiffHundredLSUI}{\ensuremath{-}0.06}
\newcommand{\DFLoHundredLSUI}{\ensuremath{-}0.27}
\newcommand{\DFHiHundredLSUI}{+0.16}

\newcommand{\DFDiffHundredAUIED}{+0.06}
\newcommand{\DFLoHundredAUIED}{\ensuremath{-}0.53}
\newcommand{\DFHiHundredAUIED}{+0.65}

\newcommand{\GNDiffTenUIEB}{\ensuremath{-}0.11}

\newcommand{\FGDiffTenUIEB}{+0.76}

\newcommand{\GNDiffTenLSUI}{\ensuremath{-}0.65}
\newcommand{\GNLoTenLSUI}{\ensuremath{-}1.29}
\newcommand{\GNHiTenLSUI}{\ensuremath{-}0.01}

\newcommand{\FGDiffTenLSUI}{+0.82}
\newcommand{\FGLoTenLSUI}{+0.22}
\newcommand{\FGHiTenLSUI}{+1.43}
\newcommand{\FGPTenLSUI}{0.111}

\newcommand{\GNDiffTenAUIED}{+0.45}

\newcommand{\FGDiffTenAUIED}{+1.03}
\newcommand{\FGLoTenAUIED}{+0.28}
\newcommand{\FGHiTenAUIED}{+1.77}

\newcommand{\FGDiffTwentyFiveUIEB}{+0.67}

\newcommand{\GNDiffTwentyFiveLSUI}{\ensuremath{-}0.50}

\newcommand{\FGDiffTwentyFiveLSUI}{+0.65}

\newcommand{\FGDiffTwentyFiveAUIED}{+0.21}

\title{\textbf{Physics-Shaped, Not Physical:\\[2pt]
Auditing Physics-Guided\\[2pt]
Underwater Image Restoration}\\[6pt]
\large Multi-seed accuracy, data efficiency, and validation against measured range}
\author{Mochamad Rizal Fauzan\textsuperscript{1}\quad Irgi Surya\textsuperscript{1}\\[3pt]
\small\textsuperscript{1}College of Electrical Engineering and Computer Science\\
\small National Taipei University of Technology, Taipei, Taiwan\\[2pt]
\small Corresponding author: M.~R.~Fauzan (t114998415@ntut.org.tw)}
\date{}

\begin{document}
\maketitle

\begin{abstract}
Physics-guided networks for underwater image restoration predict transmission and backscatter and invert the image-formation model, yet rarely test whether this physics does any work. We audit one representative design, PhysGuide-Net, with four controlled experiments: five-seed ablations on UIEB with zero-shot evaluation on the complete LSUI (4279 pairs) and AUIED3K (3000 pairs) sets; a data-efficiency study with a capacity-matched control that removes the formation model; and a validation of the predicted maps against measured range on Sea-thru. With all 800 training images the physics module gives no zero-shot gain (mean PSNR differences of at most 0.16\,dB; all 95\,\% intervals include zero) at $3.4\times$ the FLOPs. With 80 images it helps more ($+0.65$\,dB on UIEB, $+1.48$\,dB on AUIED3K; seed-to-seed s.d.\ on AUIED3K from 1.3 to 0.5\,dB), although not individually significant after correction, and the control does not reproduce this (0.8--1.0\,dB lower). Yet the predicted transmission does not behave like a physical quantity: its correlation with range is no stronger than for the range map of a different image (pooled difference $+0.02$, versus $-0.06$ for the dark-channel prior), and a Beer--Lambert fit explains 0.4\,\% of its variance. The effect of the module also changes sign with degradation severity between datasets. The benefit thus comes from a physics-shaped inductive bias, not from physics. We further show that single-run ablations can reverse sign and document implementation defects that invalidate loss ablations.
\end{abstract}

\section{Introduction}
\label{sec:intro}
Every pixel of an underwater photograph records the water it crossed. Red light vanishes within a few meters, blue--green scatter veils the scene, and both effects grow with range \cite{mcglamery1980,jaffe1990,akkaynak2018}. It is therefore natural to build this physics into restoration networks: predict a transmission map and a backscatter term, invert the image-formation model, and let a network refine the result. Many such \emph{physics-guided} designs have been proposed \cite{li2021ucolor,auied3k,li2020uwcnn}, and the idea parallels the broader movement toward physics-informed learning \cite{raissi2019,karniadakis2021}. Yet two questions that decide whether the physics deserves its name are rarely asked: \emph{is it doing any work, and is it physics at all?}

The evidence offered is often thin. Gains are commonly measured with one training run on a small benchmark such as UIEB (890 pairs) \cite{li2020uieb}, as differences of a few tenths of a decibel, although seed-to-seed variance of a network trained on 800 images can be of the same order \cite{henderson2018,bouthillier2021,picard2021}. The split that selects the checkpoint is frequently the one used for reporting. To our knowledge, predicted maps are rarely compared with measured range, even though datasets that allow this exist \cite{akkaynak2019}. And the most natural benefit of a prior, helping when data are scarce \cite{sun2017}, is seldom tested.

We therefore audit one representative design, PhysGuide-Net \cite{fauzan2026}, with four controlled experiments whose design is fixed in Section~\ref{sec:protocol}: \textbf{RQ1} a five-seed ablation on UIEB with zero-shot evaluation on the complete LSUI \cite{peng2023} and AUIED3K \cite{auied3k} sets; \textbf{RQ2} a data-efficiency study with a capacity-matched control that keeps the architecture but removes the formation model; \textbf{RQ3} a test of the predicted maps against the measured range of Sea-thru \cite{akkaynak2019}; and \textbf{RQ4} an analysis of how the effect depends on degradation severity.

\paragraph{What we find.}
\begin{itemize}\setlength\itemsep{1pt}
\item With the full training set the physics module gives \emph{no} zero-shot accuracy benefit, at $3.4\times$ the FLOPs; the only difference that survives correction lies on the checkpoint-selection split.
\item The module does help when data are scarce: the advantage grows as the training set shrinks (up to $+1.5$\,dB on AUIED3K at 80 images), and on AUIED3K the seed-to-seed spread falls from 1.3 to 0.5\,dB at that size. A control with identical capacity but no formation model does \emph{not} reproduce the gain. With five seeds these low-data gains are a consistent trend, not individually significant after correction.
\item The predicted transmission is nevertheless \emph{not} a physical quantity: on Sea-thru it follows measured range no better than the range map of a different image does, and a classical dark-channel prior does better.
\item The effect of the module depends on degradation severity, with opposite sign on LSUI and AUIED3K.
\end{itemize}
In short, the physics in this design acts as an inductive bias that is \emph{physics-shaped, not physical}: it can be useful, but not for the reason it is usually given.

\paragraph{Relation to our earlier work.} This paper extends the study that introduced PhysGuide-Net \cite{fauzan2026}, which evaluated a single seed on UIEB and 400 LSUI images. Here we add multi-seed statistics, zero-shot evaluation on complete external sets, the data-efficiency study with a control, the physical validation, and the severity analysis, and we document the implementation defects that must be excluded before such ablations are meaningful (Section~\ref{sec:repro}).

\section{Related work}
\label{sec:related}

\paragraph{Image formation and classical restoration.}
The radiance that reaches an underwater camera is commonly modeled as a direct component, attenuated with range and wavelength, plus a backscattered veiling-light component \cite{mcglamery1980,jaffe1990}; the model is formally analogous to the atmospheric scattering model of dehazing \cite{he2011}. Classical restoration estimates its unknowns with hand-crafted priors: wavelength compensation and dehazing \cite{chiang2012}, red-channel and dark-channel variants \cite{galdran2015,carlevaris2010}, blurriness- and absorption-based depth cues \cite{peng2017,drews2016}, and haze-lines with a quantitative color-chart dataset \cite{berman2021}. Fusion-based enhancement \cite{ancuti2018} avoids explicit physical modeling altogether. Akkaynak and Treibitz showed that the attenuation coefficients are not constant but depend on range and scene \cite{akkaynak2018}, which motivated Sea-thru, a method that recovers colors from RGB-D data \cite{akkaynak2019} and whose data we use here to test physical consistency.

\paragraph{Learning-based enhancement.}
Deep models trained on paired benchmarks dominate recent work. Following the UIEB benchmark and Water-Net \cite{li2020uieb}, adversarial approaches \cite{li2018emerging,fabbri2018,guo2020,islam2020}, scene-prior CNNs \cite{li2020uwcnn}, transmission-guided multi-color-space embedding \cite{li2021ucolor}, contrastive and semi-supervised training \cite{liu2022tacl,huang2023} and transformers \cite{peng2023} have been proposed; surveys are given in \cite{anwar2020,wang2019review}. Larger and more varied data have been introduced through the LSUI set \cite{peng2023}, the real-world RUIE benchmark \cite{liu2020ruie} and AUIED3K \cite{auied3k}. The authors of AUIED3K accompany it with a network that predicts a transmission map and background light. Most of these works report one trained model per method.

\paragraph{Physics-informed learning.}
Embedding physical structure into neural networks has become a general strategy of scientific machine learning \cite{raissi2019,karniadakis2021}: the physics supplies an inductive bias, which is expected to help most when data are scarce or the test distribution shifts. In underwater restoration the analog is to predict transmission and backscatter and to condition or regularize the restoration branch on them. Whether the predicted quantities are physically meaningful, and whether the bias pays off, is rarely tested directly; this is the question we pursue.

\paragraph{Evaluation practice.}
Differences of a few tenths of a decibel are routinely reported although seed-to-seed variance can be of the same order \cite{henderson2018,bouthillier2021,picard2021}; calls for multi-run reporting and careful statistics \cite{dodge2019,pineau2021,demsar2006} have had limited uptake in enhancement papers. Full-reference scores such as the peak signal-to-noise ratio (PSNR) and the structural similarity (SSIM) \cite{wang2004} are complemented by non-reference measures such as the underwater image quality measure (UIQM) \cite{panetta2016} and the underwater color image quality evaluation (UCIQE) \cite{yang2015}; color accuracy against charts is conventionally expressed as an angular error \cite{hordley2004,gijsenij2011}. Dataset shift \cite{bendavid2010} and the amount of training data \cite{sun2017} are known determinants of generalization, which is why we examine both.

\section{PhysGuide-Net}
\label{sec:method}

\begin{figure}[t]
\centering
\includegraphics[width=\textwidth]{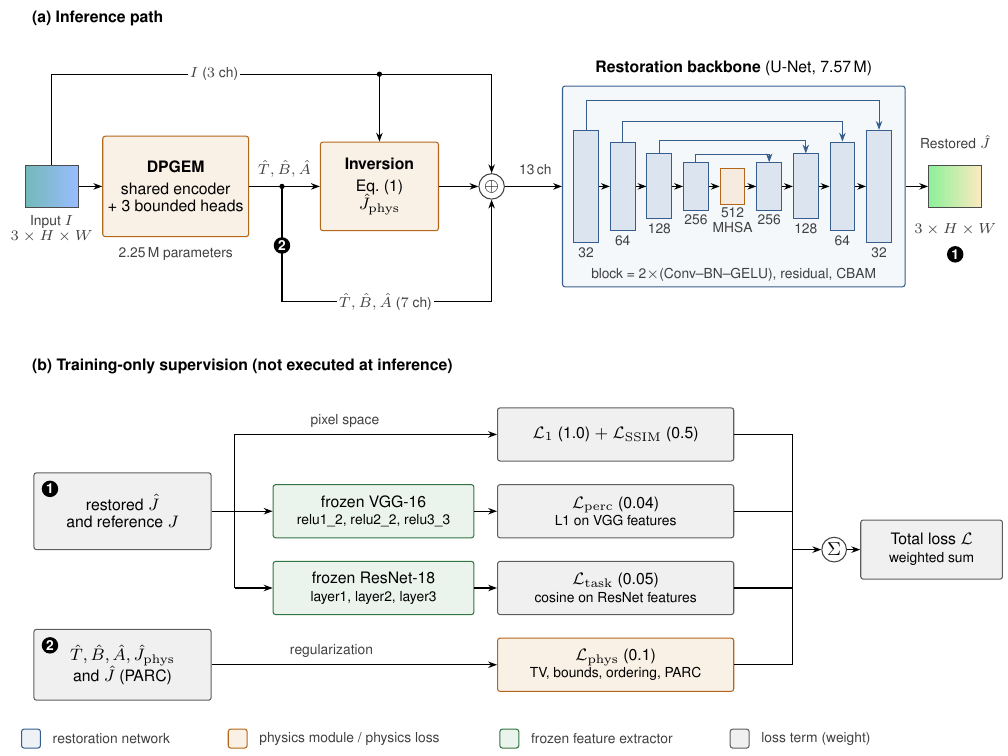}
\caption{System architecture of PhysGuide-Net. \textbf{(a) Inference path:} the degraded image $I$ passes through the DPGEM, which predicts the transmission, backscatter and attenuation maps; the formation-model inversion of Eq.~(\ref{eq:phys}) turns them into the physics-recovered prior $\hat J_{\mathrm{phys}}$; $I$, the three maps and $\hat J_{\mathrm{phys}}$ (13 channels) are concatenated and restored by a U-Net with CBAM attention and a self-attention bottleneck (MHSA). \textbf{(b) Training-only supervision:} the restored image is compared with the reference in pixel space and in the feature spaces of a frozen VGG-16 (perceptual) and a frozen ResNet-18 (task preserving), while the physics maps are regularized; markers 1 and 2 link the two panels. Numbers in parentheses are loss weights.}
\label{fig:arch}
\end{figure}

PhysGuide-Net \cite{fauzan2026} couples a lightweight estimator of the underwater image-formation variables with a restoration backbone that is \emph{conditioned} on them (Fig.~\ref{fig:arch}). Three design choices define the system: (i) physical variables are predicted per pixel and turned into an explicit physics-recovered prior; (ii) the restoration network consumes the raw image together with all physical maps; (iii) training is supervised both in pixel space and in the feature space of frozen networks, while the extra supervision is discarded at inference.

\subsection{Degradation-aware physics-guided estimation}
Given $I\in[0,1]^{3\times H\times W}$, the degradation-aware physics-guided estimation module (DPGEM) uses a shared convolutional encoder that feeds three heads with bounded outputs: $\hat T=\sigma(\cdot)\in(0,1)$ (one channel), $\hat B=0.5\,\sigma(\cdot)\in(0,0.5)$ and $\hat A=2\,\sigma(\cdot)\in(0,2)$ (three channels each). Inverting the simplified formation model $I=J\,T+B\,(1-T)$ gives the physics-recovered prior
\begin{equation}
\hat J_{\mathrm{phys}}=\operatorname{clip}_{[0,1]}\!\left(\frac{I-\hat B\,(1-\hat T)}{\max(\hat T,\epsilon)}\right).
\label{eq:phys}
\end{equation}
The module has 2.25~M parameters and no direct supervision: $\hat T,\hat B,\hat A$ are learned only through the reconstruction objective and the regularizers of Section~\ref{sec:loss}.

\subsection{Physics-conditioned restoration backbone}
The 13-channel tensor $[I,\hat T,\hat B,\hat A,\hat J_{\mathrm{phys}}]$ is processed by a four-level U-Net with encoder widths $32,64,128,256$, a 512-channel bottleneck and a mirrored decoder with transposed-convolution upsampling. Each block consists of two Conv--BN--GELU layers with a residual shortcut, dropout of 0.1 and a convolutional block attention module (CBAM) \cite{woo2018}; a compact four-head self-attention layer (MHSA) on an $8\times8$ token grid at the bottleneck models global color statistics. The backbone has 7.57~M parameters; the frozen ImageNet ResNet-18 \cite{he2016} used for the task loss is excluded from the 9.82~M trainable inference parameters because it is never executed at test time.

\subsection{Training objective}
\label{sec:loss}
With $\hat J$ the restored image and $J$ the reference,
\begin{equation}
\mathcal{L}=\mathcal{L}_{1}+0.5\,\mathcal{L}_{\mathrm{SSIM}}+0.04\,\mathcal{L}_{\mathrm{perc}}+0.1\,\mathcal{L}_{\mathrm{phys}}+0.05\,\mathcal{L}_{\mathrm{task}}.
\end{equation}
$\mathcal{L}_{\mathrm{SSIM}}=1-\mathrm{SSIM}$ \cite{wang2004}. $\mathcal{L}_{\mathrm{perc}}$ is the mean absolute difference of normalized VGG-16 \cite{simonyan2015} features at relu1\_2, relu2\_2 and relu3\_3 \cite{johnson2016}. $\mathcal{L}_{\mathrm{task}}=\frac{1}{|\mathcal{S}|}\sum_{l\in\mathcal{S}}\big\langle 1-\cos\!\big(\varphi_l(\hat J),\varphi_l(J)\big)\big\rangle$ averages one minus the cosine similarity over spatial positions of the ResNet-18 feature maps $\varphi_l$, $\mathcal{S}=\{\text{layer1},\text{layer2},\text{layer3}\}$, with the reference features detached. $\mathcal{L}_{\mathrm{phys}}$ combines total variation (TV) on $\hat T$, penalties for $\hat T\notin[0.05,1]$ and $\hat B\notin[0,0.5]$, a wavelength-ordering penalty $\max(0,\hat A_G-\hat A_R)+\max(0,\hat A_B-\hat A_G)$ and an $\ell_1$ term between $\hat J$ and the detached $\hat J_{\mathrm{phys}}$ (physics-auxiliary reconstruction consistency, PARC). Because $\hat B$ is sigmoid-bounded, its bound penalty is identically zero; the remaining terms are active.

\subsection{Controlled variants}
\label{sec:variants}
All variants share data, split, optimizer and schedule and differ in a single component:
\begin{itemize}\setlength\itemsep{1pt}
\item \textbf{Full}: as described.
\item \textbf{w/o physics module}: DPGEM and physics fusion removed (the backbone sees only RGB), $\mathcal{L}_{\mathrm{phys}}$ disabled; $\mathcal{L}_{\mathrm{task}}$ and $\mathcal{L}_{\mathrm{perc}}$ kept.
\item \textbf{w/o task loss}: $\mathcal{L}_{\mathrm{task}}$ disabled.
\item \textbf{Plain U-Net}: the same backbone trained with $\mathcal{L}_{1}$, $\mathcal{L}_{\mathrm{SSIM}}$ and $\mathcal{L}_{\mathrm{perc}}$ only; against ``w/o physics module'' it isolates $\mathcal{L}_{\mathrm{task}}$.
\end{itemize}

\section{Study design}
\label{sec:protocol}

We organize the study around four research questions, each answered by a dedicated experiment:
\begin{description}\setlength\itemsep{1pt}
\item[RQ1 (accuracy).] With everything else controlled, does the physics guidance improve restoration, in domain and zero-shot?
\item[RQ2 (data efficiency).] Does the prior help when the amount of training data shrinks?
\item[RQ3 (physical validity).] Do the predicted maps behave like physical quantities, i.e.\ does $\hat T$ fall with measured range?
\item[RQ4 (severity).] Does the effect depend on how strongly an image is degraded?
\end{description}

\subsection{Data}
\textbf{UIEB} (Underwater Image Enhancement Benchmark) \cite{li2020uieb} provides 890 raw/reference pairs; we keep the fixed 800/90 train/validation split of the earlier study. Two sets are used only for zero-shot evaluation (never for training or checkpoint selection), on their \emph{complete} content: \textbf{LSUI} (Large-Scale Underwater Image dataset) \cite{peng2023} (\NImgLSUI{} pairs in the copy available to us; the original article reports 5004) and \textbf{AUIED3K} (Andaman Underwater Image Enhancement Dataset) \cite{auied3k} (\NImgAUIED{} pairs, all $3840\times2160$). In AUIED3K the references are real clear photographs taken at depths of up to 5\,m, whereas the raw images are \emph{synthesized} from them with the dataset authors' degradation algorithm, which varies per-channel absorption and noise from clear to turbid water \cite{auied3k}; the set therefore tests transfer from real to synthetic degradation. For \textbf{Sea-thru} \cite{akkaynak2019} we use the linear RGB images and measured range maps of five sites (D1--D5); it has no restoration reference and is used only for RQ3. All images are resized to $256\times256$ (bicubic). For comparison with the earlier study, the 400-image LSUI subset used there (``LSUI-400'') is also reported.

\subsection{Training and evaluation}
AdamW \cite{loshchilov2019} (learning rate $2\times10^{-4}$, weight decay $10^{-4}$), cosine annealing with warm restarts ($T_0=50$), batch size 8, mixed precision, gradient clipping at 1.0, at most 100 epochs with early stopping (patience 15) on validation PSNR. Augmentation: horizontal/vertical flips, $90^\circ$ rotations and input-only color jitter. Seeds 42--46 control initialization and augmentation order; the split is identical for every run. PSNR and SSIM \cite{wang2004} are computed on $[0,1]$ RGB at $256\times256$ for the epoch with the best validation PSNR. Since UIEB-val selects the checkpoint, its numbers are optimistic and are marked $^\dagger$; the zero-shot sets are the primary evidence.

\subsection{Statistics}
We report the mean $\pm$ standard deviation over five seeds and paired differences between variants, using (i) a paired $t$-test with a 95\,\% confidence interval (CI) on the per-seed means, and (ii) a Wilcoxon signed-rank test \cite{wilcoxon1945} on seed-averaged per-image scores. $p$-values are Holm-corrected \cite{holm1979} within each experiment: for RQ1 over 4 comparisons $\times$ 4 evaluation sets $\times$ 2 metrics; for RQ2 over the 12 paired gains and, separately, over the six comparisons of each control contrast. The seed-level test has few degrees of freedom but reflects training variance; the image-level test is more sensitive but conditions on the trained models. We therefore base conclusions on effect sizes and seed-level intervals.

\subsection{RQ2: data-efficiency design}
Nested random subsets of the 800 training images (10\,\%, 25\,\%, 50\,\% $=$ 80, 200, 400 images) are drawn once from a fixed permutation, so that every subset contains the smaller ones and is identical across seeds and variants. Full and ``w/o physics module'' are trained on each subset with seeds 42--46 under the unchanged protocol (same epochs, same early stopping); validation uses the same 90 images. To keep the evaluation of each run under a minute, the zero-shot sets are represented by fixed random subsets of 1000 images each; the 100\,\% runs of the main study are re-scored on the same subsets.

\subsection{RQ3: physical validation on Sea-thru}
Sea-thru images are linear, so each image is exposure-normalized (99.5th percentile) and sRGB-encoded before it is fed to the networks, which were trained on display-referred images; this domain shift is a caveat. We use an evenly spaced subset of at most 100 images per site (\emph{\STNimg{}} images in total) and the measured range maps (invalid pixels masked). A physical transmission decreases with range, so we compute per image the Spearman correlation $\rho(\hat T,d)$ between $\hat T$ and the range $d$ (expected $<0$). Because spatial layout alone (for example, far regions near the top of the frame) can induce a correlation, every statistic is compared with a \emph{null control} that correlates $\hat T$ with the range map of a different image of the same site; the image-specific signal is $\Delta=\rho(\hat T,d)-\rho(\hat T,d_{\mathrm{null}})$ and should be negative. For reference we apply the same protocol to a classical dark-channel-prior (DCP) transmission \cite{he2011} (patch $15\times15$, $\omega=0.95$, background light from the brightest 0.1\,\% of the dark-channel pixels). Further checks are $\rho(\hat B,d)$ (expected $>0$), the $R^2$ of a log-linear fit $\log\hat T\sim d$ (Beer--Lambert), and the range dependence of the red-to-green chromaticity ratio of the restored images in linear light, which a range-invariant restoration should reduce. Per-image values are averaged over the five seeds before aggregation.

\subsection{RQ4: severity analysis}
Images of each split are binned into fifths by the PSNR of the unprocessed input against its reference. Within each bin we report the seed-averaged paired difference with a bootstrap 95\,\% interval over images, and the Spearman correlation between input PSNR and the difference. This analysis was added after seeing the main results and is exploratory.

\section{Results}
\label{sec:results}

\subsection{RQ1: accuracy in domain and zero-shot}
\label{sec:rq1}
\begin{figure}[t]
\centering
\includegraphics[width=\linewidth]{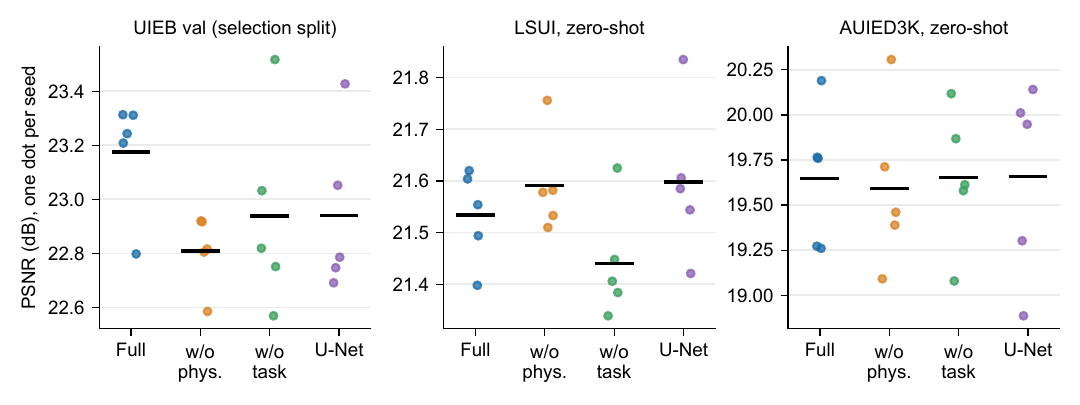}
\caption{PSNR of every variant, one dot per seed; the black bar is the mean over seeds. The UIEB-val panel is the checkpoint-selection split.}
\label{fig:variants}
\end{figure}

Table~\ref{tab:main} and Fig.~\ref{fig:variants} summarize the five seeds per variant. On UIEB-val, which also selects the checkpoint, the full model has the highest mean PSNR (\MeanUIEBFullPSNR\,dB) and the variant without the physics module the lowest (\MeanUIEBNoPhysPSNR\,dB). On the two zero-shot sets the picture is flat: the variant means span \MeanLSUINoTaskPSNR--\MeanLSUIUnetPSNR\,dB on LSUI and \MeanAUIEDNoPhysPSNR--\MeanAUIEDUnetPSNR\,dB on AUIED3K, and the full model is the best variant on neither (the plain U-Net is). The seed-to-seed standard deviation (0.09--0.15\,dB on LSUI, 0.39--0.54\,dB on AUIED3K) is of the same order as the gaps between variants; SSIM shows the same pattern (Table~\ref{tab:main}, lower block).

\begin{table}[!htbp]
\centering\small
\caption{Mean $\pm$ standard deviation over five seeds. Params: trainable inference parameters (frozen task encoder excluded); GFLOPs at $256\times256$. Bold: best mean per column. $^\dagger$\,UIEB-val also selects the checkpoint (optimistic).}
\label{tab:main}
\resizebox{\linewidth}{!}{\begin{tabular}{@{}l c c c c c @{}}
\toprule
& & & \multicolumn{3}{c}{PSNR (dB) $\uparrow$} \\
\cmidrule(lr){4-6}
Variant & Params (M) & GFLOPs & UIEB val ($n{=}90$)$^\dagger$ & LSUI ($n{=}4279$) & AUIED3K ($n{=}3000$) \\
\midrule
PhysGuide-Net (full) & 9.82 & 32.2 & \textbf{23.17 $\pm$ 0.22} & 21.53 $\pm$ 0.09 & 19.65 $\pm$ 0.39 \\
w/o physics module & 7.57 & 9.6 & 22.81 $\pm$ 0.14 & 21.59 $\pm$ 0.10 & 19.59 $\pm$ 0.46 \\
w/o task loss & 9.82 & 32.2 & 22.94 $\pm$ 0.36 & 21.44 $\pm$ 0.11 & 19.65 $\pm$ 0.39 \\
Plain U-Net & 7.57 & 9.6 & 22.94 $\pm$ 0.31 & \textbf{21.60 $\pm$ 0.15} & \textbf{19.66 $\pm$ 0.54} \\
\midrule
& & & \multicolumn{3}{c}{SSIM $\uparrow$} \\
\cmidrule(lr){4-6}
PhysGuide-Net (full) &  &  & \textbf{0.906 $\pm$ 0.002} & 0.858 $\pm$ 0.002 & 0.907 $\pm$ 0.005 \\
w/o physics module &  &  & 0.900 $\pm$ 0.002 & 0.858 $\pm$ 0.003 & \textbf{0.913 $\pm$ 0.008} \\
w/o task loss &  &  & 0.902 $\pm$ 0.004 & 0.856 $\pm$ 0.001 & 0.911 $\pm$ 0.005 \\
Plain U-Net &  &  & 0.903 $\pm$ 0.005 & \textbf{0.859 $\pm$ 0.003} & 0.913 $\pm$ 0.008 \\
\bottomrule
\end{tabular}
}
\end{table}

\begin{figure}[t]
\centering
\includegraphics[width=0.82\linewidth]{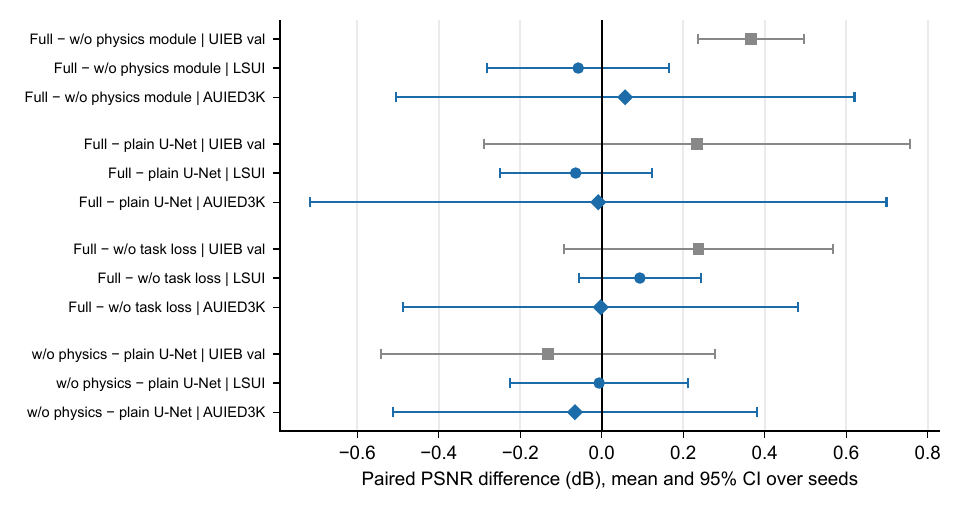}
\caption{Paired PSNR differences between variants (mean and 95\,\% CI over five seeds). Gray: checkpoint-selection split; blue: zero-shot sets.}
\label{fig:forest}
\end{figure}

\paragraph{Paired comparisons.} Table~\ref{tab:paired} and Fig.~\ref{fig:forest} give paired differences. Only one difference survives Holm correction at the seed level: the full model is ahead of the variant without the physics module by \DiffUIEBFullVsNoPhysPSNR\,dB on UIEB-val (95\,\% CI [\LoUIEBFullVsNoPhysPSNR, \HiUIEBFullVsNoPhysPSNR], $p=\PsUIEBFullVsNoPhysPSNR$). That is the checkpoint-selection split, and the effect does not transfer: on LSUI the difference is \DiffLSUIFullVsNoPhysPSNR\,dB [\LoLSUIFullVsNoPhysPSNR, \HiLSUIFullVsNoPhysPSNR] and on AUIED3K \DiffAUIEDFullVsNoPhysPSNR\,dB [\LoAUIEDFullVsNoPhysPSNR, \HiAUIEDFullVsNoPhysPSNR]. Against the plain U-Net, the relevant baseline for any claim that physics guidance helps, the full model differs by \DiffLSUIFullVsUnetPSNR\,dB on LSUI [\LoLSUIFullVsUnetPSNR, \HiLSUIFullVsUnetPSNR] and \DiffAUIEDFullVsUnetPSNR\,dB on AUIED3K [\LoAUIEDFullVsUnetPSNR, \HiAUIEDFullVsUnetPSNR]. The LSUI intervals exclude differences larger than about 0.3\,dB in either direction; the AUIED3K intervals are wider (up to $\pm0.7$\,dB) because AUIED3K scores vary by 0.4--0.5\,dB across seeds. The task-preserving loss shows no measurable effect either: \DiffLSUIFullVsNoTaskPSNR\,dB on LSUI and \DiffAUIEDFullVsNoTaskPSNR\,dB on AUIED3K. At the image level the large number of images makes even differences of 0.06\,dB statistically detectable (Wilcoxon $p<0.001$ for several comparisons, $|d_z|\le0.19$); because that test ignores training variance we treat such differences as detectable but practically negligible.

\begin{table}[!htbp]
\centering\small
\caption{Paired PSNR differences (dB). $\Delta$: mean difference of per-seed means with 95\,\% CI and Holm-corrected paired $t$ $p$-value over five seeds; $\Delta_{\text{img}}$: difference of seed-averaged per-image PSNR with Holm-corrected Wilcoxon $p$-value. $^\dagger$\,checkpoint-selection split.}
\label{tab:paired}
\resizebox{\linewidth}{!}{\begin{tabular}{@{}l l r c c r c@{}}
\toprule
Comparison & Split & $\Delta$PSNR & 95\% CI (seeds) & $p_{\text{seed}}$ & $\Delta$PSNR$_{\text{img}}$ & $p_{\text{img}}$ \\
\midrule
Full $-$ w/o physics module & UIEB val ($n{=}90$)$^\dagger$ & +0.37 & [+0.24, +0.50] & 0.046 & +0.37 & 0.006 \\
 & LSUI ($n{=}4279$) & \ensuremath{-}0.06 & [\ensuremath{-}0.28, +0.17] & 1.000 & \ensuremath{-}0.06 & $<$0.001 \\
 & AUIED3K ($n{=}3000$) & +0.06 & [\ensuremath{-}0.51, +0.62] & 1.000 & +0.06 & $<$0.001 \\
\addlinespace
Full $-$ plain U-Net & UIEB val ($n{=}90$)$^\dagger$ & +0.23 & [\ensuremath{-}0.29, +0.76] & 1.000 & +0.23 & 0.118 \\
 & LSUI ($n{=}4279$) & \ensuremath{-}0.06 & [\ensuremath{-}0.25, +0.12] & 1.000 & \ensuremath{-}0.06 & $<$0.001 \\
 & AUIED3K ($n{=}3000$) & \ensuremath{-}0.01 & [\ensuremath{-}0.72, +0.70] & 1.000 & \ensuremath{-}0.01 & 1.000 \\
\addlinespace
Full $-$ w/o task loss & UIEB val ($n{=}90$)$^\dagger$ & +0.24 & [\ensuremath{-}0.09, +0.57] & 1.000 & +0.24 & 0.006 \\
 & LSUI ($n{=}4279$) & +0.09 & [\ensuremath{-}0.06, +0.24] & 1.000 & +0.09 & $<$0.001 \\
 & AUIED3K ($n{=}3000$) & 0.00 & [\ensuremath{-}0.49, +0.48] & 1.000 & 0.00 & 0.274 \\
\addlinespace
w/o physics $-$ plain U-Net & UIEB val ($n{=}90$)$^\dagger$ & \ensuremath{-}0.13 & [\ensuremath{-}0.54, +0.28] & 1.000 & \ensuremath{-}0.13 & 0.304 \\
 & LSUI ($n{=}4279$) & \ensuremath{-}0.01 & [\ensuremath{-}0.22, +0.21] & 1.000 & \ensuremath{-}0.01 & $<$0.001 \\
 & AUIED3K ($n{=}3000$) & \ensuremath{-}0.07 & [\ensuremath{-}0.51, +0.38] & 1.000 & \ensuremath{-}0.07 & $<$0.001 \\
\bottomrule
\end{tabular}
}
\end{table}

\paragraph{Single runs mislead.} In the single-seed run of the earlier study, removing the task loss appeared to \emph{improve} PSNR on the 400-image LSUI subset by 0.43\,dB (image-level $p<0.001$). With five seeds the full model is instead ahead of the variant without the task loss by \DiffLSUIsubFullVsNoTaskPSNR\,dB (95\,\% CI [\LoLSUIsubFullVsNoTaskPSNR, \HiLSUIsubFullVsNoTaskPSNR]). The image-level test of the single run was ``significant'' because it described one trained model, not the training procedure.

\subsection{RQ2: data efficiency and a capacity-matched control}
\label{sec:rq2}
\begin{figure}[t]
\centering
\includegraphics[width=\linewidth]{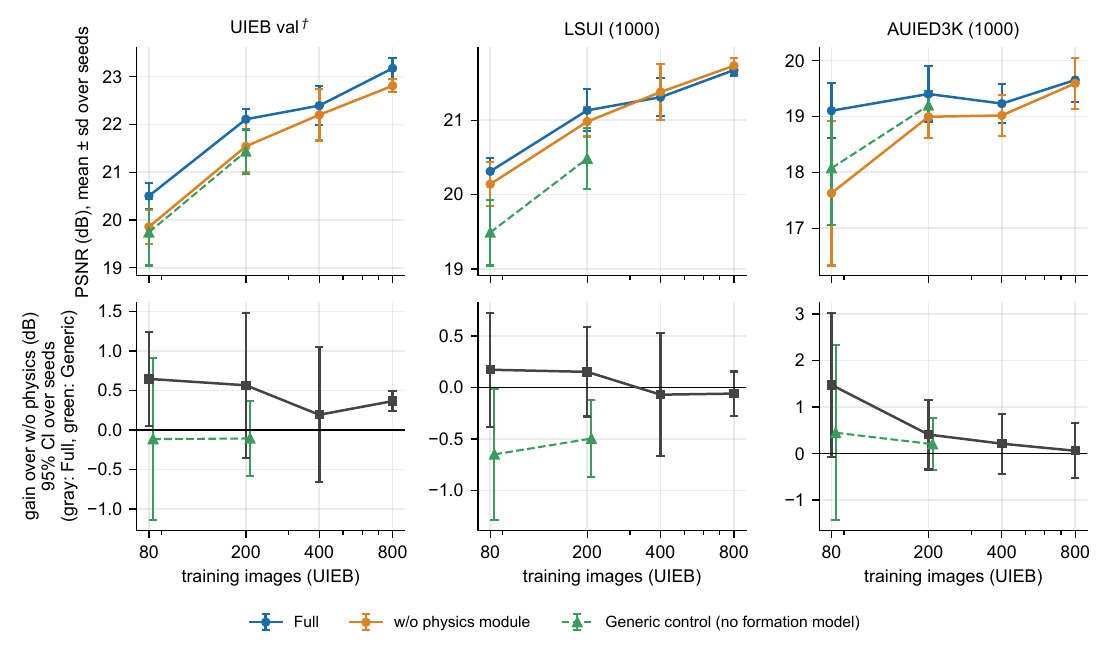}
\caption{Data-efficiency curves. Top: mean PSNR $\pm$ s.d. over five seeds for the full model, the variant without the physics module and (at 80 and 200 images) the capacity-matched generic control. Bottom: paired gain over the variant without the module with 95\,\% CI (gray: full model; green: generic control).}
\label{fig:datafrac}
\end{figure}

Fig.~\ref{fig:datafrac} and Table~\ref{tab:datafrac} show how the benefit of the physics module changes with the amount of training data. Two observations stand out.

\paragraph{The advantage grows as data shrink.} With all 800 images the paired gain over the variant without the module is \DFDiffHundredLSUI\,dB on LSUI and \DFDiffHundredAUIED\,dB on AUIED3K. With 80 images (10\,\%) it is \DFDiffTenUIEB\,dB on UIEB-val (95\,\% CI [\DFLoTenUIEB, \DFHiTenUIEB]) and \DFDiffTenAUIED\,dB on AUIED3K ([\DFLoTenAUIED, \DFHiTenAUIED]), and on AUIED3K it decays monotonically with the training-set size (\DFDiffTenAUIED, \DFDiffTwentyFiveAUIED, \DFDiffFiftyAUIED, \DFDiffHundredAUIED\,dB for 80, 200, 400, 800 images). On LSUI the gain stays within $\pm0.2$\,dB. At 80 images the seed-to-seed standard deviation on AUIED3K is also lower with the module (0.5\,dB against 1.3\,dB without it), but this is not a consistent pattern across splits and fractions (at 200 images it is 0.5 against 0.4\,dB). These low-data gains are a consistent trend rather than established effects: at 10\,\% only the UIEB-val interval excludes zero, and after Holm correction over the twelve comparisons of this experiment only the full-data UIEB-val difference remains significant ($p=\DFPHundredUIEB$).

\paragraph{Capacity alone does not reproduce it.} The ``generic'' control keeps the encoder, decoder and the $1/3/3/3$ output channels of the module (9.83\,M parameters, $+0.09\,\%$) but replaces the formation-model inversion by an ordinary head and drops the physics losses. At 80 images the full model is better than this control by \FGDiffTenUIEB\,dB on UIEB-val, \FGDiffTenLSUI\,dB on LSUI (CI [\FGLoTenLSUI, \FGHiTenLSUI]) and \FGDiffTenAUIED\,dB on AUIED3K ([\FGLoTenAUIED, \FGHiTenAUIED]); at 200 images the differences are \FGDiffTwentyFiveUIEB, \FGDiffTwentyFiveLSUI{} and \FGDiffTwentyFiveAUIED\,dB. The generic variant is not better than having no module at all (generic minus w/o module: \GNDiffTenUIEB, \GNDiffTenLSUI{} and \GNDiffTenAUIED\,dB at 80 images), and on LSUI it is worse than having no module (\GNDiffTenLSUI\,dB, [\GNLoTenLSUI, \GNHiTenLSUI], at 80 images; \GNDiffTwentyFiveLSUI\,dB at 200 images). None of the six full-versus-generic comparisons is individually significant after Holm correction (smallest $p=\FGPTenLSUI$), so this too is a directional result obtained with five seeds. The control removes the inversion and the regularizers together, so their contributions cannot be separated with this design.

\begin{table}[!htbp]
\centering\small
\caption{Paired PSNR gain of the full model over the variant without the physics module (dB; mean and 95\,\% CI over five seeds) as a function of the number of UIEB training images. UIEB-val: 90 images ($^\dagger$ selection split); LSUI and AUIED3K: fixed random subsets of 1000 images each.}
\label{tab:datafrac}
\resizebox{\linewidth}{!}{%
\begin{tabular}{@{}r l l l@{}}
\toprule
Training images & UIEB-val$^\dagger$ & LSUI & AUIED3K \\
\midrule
80 (10\,\%)  & \DFDiffTenUIEB{} [\DFLoTenUIEB, \DFHiTenUIEB] & \DFDiffTenLSUI{} [\DFLoTenLSUI, \DFHiTenLSUI] & \DFDiffTenAUIED{} [\DFLoTenAUIED, \DFHiTenAUIED] \\
200 (25\,\%) & \DFDiffTwentyFiveUIEB{} [\DFLoTwentyFiveUIEB, \DFHiTwentyFiveUIEB] & \DFDiffTwentyFiveLSUI{} [\DFLoTwentyFiveLSUI, \DFHiTwentyFiveLSUI] & \DFDiffTwentyFiveAUIED{} [\DFLoTwentyFiveAUIED, \DFHiTwentyFiveAUIED] \\
400 (50\,\%) & \DFDiffFiftyUIEB{} [\DFLoFiftyUIEB, \DFHiFiftyUIEB] & \DFDiffFiftyLSUI{} [\DFLoFiftyLSUI, \DFHiFiftyLSUI] & \DFDiffFiftyAUIED{} [\DFLoFiftyAUIED, \DFHiFiftyAUIED] \\
800 (100\,\%) & \DFDiffHundredUIEB{} [\DFLoHundredUIEB, \DFHiHundredUIEB] & \DFDiffHundredLSUI{} [\DFLoHundredLSUI, \DFHiHundredLSUI] & \DFDiffHundredAUIED{} [\DFLoHundredAUIED, \DFHiHundredAUIED] \\
\bottomrule
\end{tabular}}
\end{table}

\subsection{RQ3: are the predicted maps physical?}
\label{sec:rq3}
\begin{figure}[t]
\centering
\includegraphics[width=\linewidth]{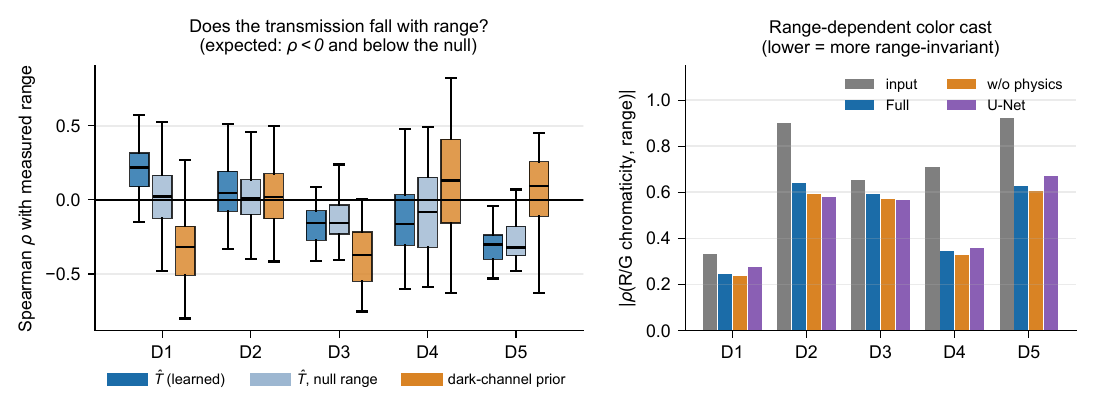}
\caption{Sea-thru validation. Left: Spearman correlation with measured range per image (boxes: quartiles over images; expected negative and below the null). Right: range dependence of the R/G chromaticity ratio of the input and of the restorations (lower is more range-invariant).}
\label{fig:seathru}
\end{figure}

Table~\ref{tab:seathru} and Fig.~\ref{fig:seathru} report the Sea-thru validation on \STNimg{} images. Pooled over the five sites, the correlation between the learned transmission and the measured range is $\rho(\hat T,d)=\STRhoT$, hardly below zero and not below the null control (\STRhoNull). The image-specific signal $\Delta=\rho(\hat T,d)-\rho(\hat T,d_{\mathrm{null}})$ is \STDeltaT, i.e.\ of the \emph{wrong} sign (Wilcoxon $p=\STWilcoxP$); a physical transmission would give a clearly negative $\Delta$. The sign of $\rho$ is physically correct for \STFracNeg\,\% of the images, against \STFracNull\,\% for the null range map and \STFracDCP\,\% for the classical dark-channel prior. Site by site, $\hat T$ is \emph{positively} correlated with range at D1 ($\Delta=+0.10$, CI above zero), indistinguishable from the null at D3--D5, whereas the dark-channel prior has a negative median $\Delta$ at four of the five sites. The backscatter estimate has the expected positive correlation with range in only \STFracBpos\,\% of the images (pooled $\rho=\STRhoB$), and a log-linear (Beer--Lambert) fit of $\hat T$ against range explains only \STRsq\,\% of the variance.

The restorations themselves reduce the range dependence of the red-to-green chromaticity ratio (median $|\rho|$ from \CastInput{} for the input to \CastFull{} for the full model), but equally so without the physics module (\CastNoPhys) and for the plain U-Net (\CastUnet): the physics guidance adds no range invariance (right panel of Fig.~\ref{fig:seathru}).

\begin{table}[!htbp]
\centering\small
\caption{Sea-thru physical validation (median over images; each image averaged over five seeds). $\rho(\hat T,d)$: Spearman correlation with measured range (physical expectation: negative); null: correlation with the range map of another image of the same site; $\Delta$: difference between the two with bootstrap 95\,\% CI (physical expectation: negative); DCP $\Delta$: the same for the dark-channel-prior transmission; $\rho(\hat B,d)$: backscatter (expected positive); $R^2_{\log\hat T}$: variance of $\log\hat T$ explained by range.}
\label{tab:seathru}
\resizebox{\linewidth}{!}{\begin{tabular}{@{}l r c c c c c c@{}}
\toprule
Site & $n$ & $\rho(\hat T,d)$ & null & $\Delta$ [95\,\% CI] & DCP $\Delta$ & $\rho(\hat B,d)$ & $R^2_{\log\hat T}$ \\
\midrule
D1 & 100 & +0.22 & +0.02 & +0.10 [+0.08, +0.17] & \ensuremath{-}0.20 & +0.19 & 0.006 \\
D2 & 100 & +0.05 & +0.01 & +0.03 [0.00, +0.05] & \ensuremath{-}0.03 & \ensuremath{-}0.03 & 0.005 \\
D3 & 68 & \ensuremath{-}0.16 & \ensuremath{-}0.16 & \ensuremath{-}0.01 [\ensuremath{-}0.03, 0.00] & \ensuremath{-}0.10 & \ensuremath{-}0.04 & 0.001 \\
D4 & 100 & \ensuremath{-}0.16 & \ensuremath{-}0.08 & \ensuremath{-}0.01 [\ensuremath{-}0.05, +0.03] & \ensuremath{-}0.06 & \ensuremath{-}0.16 & 0.005 \\
D5 & 43 & \ensuremath{-}0.30 & \ensuremath{-}0.32 & \ensuremath{-}0.03 [\ensuremath{-}0.06, +0.03] & +0.09 & \ensuremath{-}0.22 & 0.004 \\
\midrule
Pooled & 411 & \ensuremath{-}0.07 & \ensuremath{-}0.10 & +0.02 & \ensuremath{-}0.06 & \ensuremath{-}0.05 & 0.004 \\
\bottomrule
\end{tabular}
}
\end{table}

\subsection{RQ4: dependence on degradation severity}
\label{sec:rq4}
\begin{figure}[t]
\centering
\includegraphics[width=\linewidth]{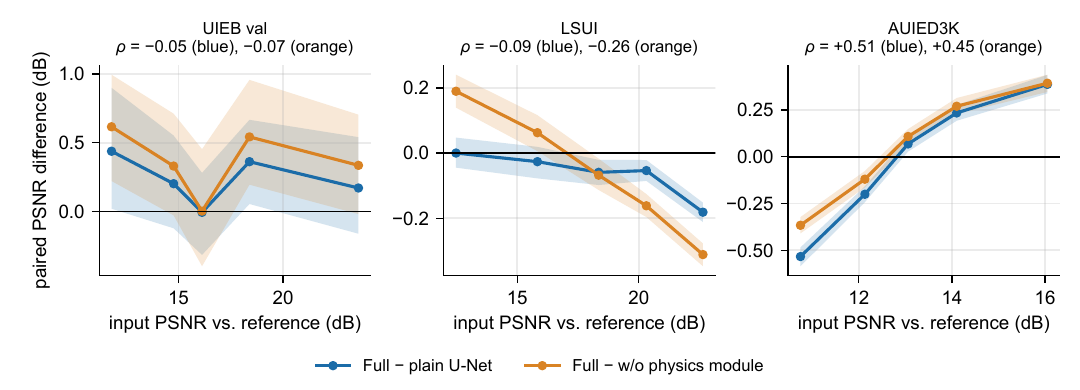}
\caption{Paired PSNR difference versus degradation severity (input PSNR against the reference, fifths of each split). Shaded: bootstrap 95\,\% CI over images; $\rho$: Spearman correlation between input PSNR and the difference.}
\label{fig:severity}
\end{figure}

Physics priors should matter most when the degradation is strong. Fig.~\ref{fig:severity} bins each split by input PSNR and shows the paired difference of the full model against the plain U-Net (blue) and against the variant without the physics module (orange). The dependence on severity is real, but \emph{its direction differs between the two zero-shot sets}.
\begin{itemize}\setlength\itemsep{1pt}
\item \textbf{UIEB-val} ($n=90$): no trend (Spearman $\rho=-0.05$ and $-0.07$); the difference is positive in most bins.
\item \textbf{LSUI}: relative to the variant without physics, the module helps the most degraded fifth (mean input PSNR 12.5\,dB: $+0.19$\,dB, 95\,\% CI $[+0.14,+0.24]$) and hurts the least degraded fifth (22.6\,dB: $-0.31$\,dB, $[-0.35,-0.28]$; $\rho=-0.26$), the direction expected of a physical prior. Against the plain U-Net the differences are small ($|\Delta|\le0.18$\,dB in every bin, $\rho=-0.09$).
\item \textbf{AUIED3K}: the pattern is reversed ($\rho=+0.51$ against the plain U-Net, $+0.45$ against w/o physics). In the most degraded fifth (10.7\,dB) the physics-guided model is $0.53$\,dB \emph{worse} than the plain U-Net ($[-0.59,-0.48]$; $-0.37$\,dB against w/o physics), and in the least degraded fifth (16.0\,dB) it is $0.39$\,dB better ($[+0.34,+0.44]$).
\end{itemize}
The overall mean differences thus hide effects of both signs. Since the same trained models are evaluated zero-shot on both sets, a consistent advantage of physics guidance under strong degradation is not supported; the data indicate an interaction between the module and the dataset. These intervals are conditional on the trained models and exclude training variance. Input PSNR is also a crude proxy for physical degradation: on AUIED3K it mainly reflects the strength of the dataset authors' synthetic degradation (10.7--16\,dB across the set).

\subsection{Qualitative examples and cost}
\begin{figure}[p]
\centering
\includegraphics[width=0.9\linewidth]{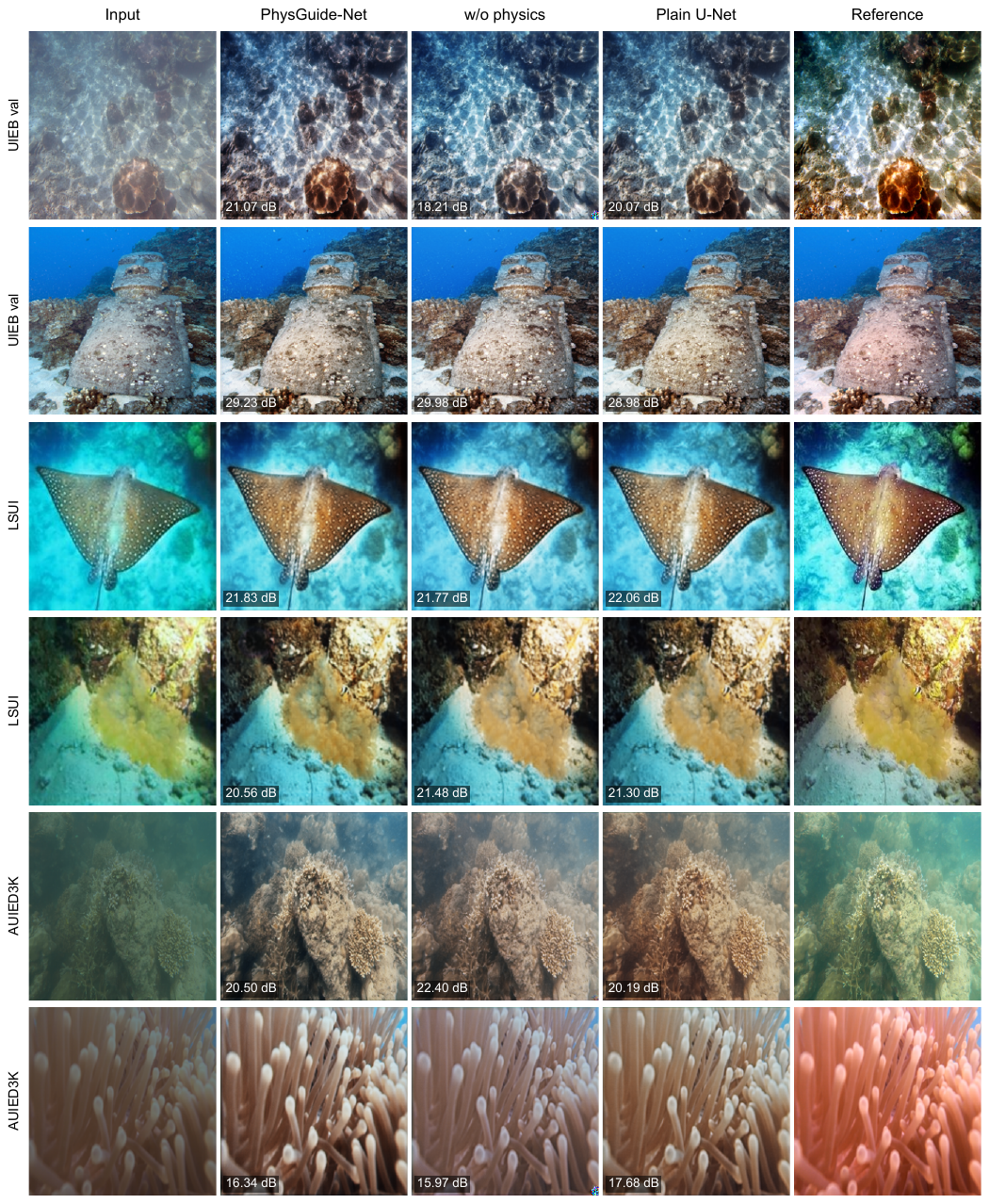}
\caption{Random examples (two per dataset). Columns: input, PhysGuide-Net, w/o physics module, plain U-Net, reference; the PSNR of each restoration is annotated.}
\label{fig:qual}
\end{figure}

Fig.~\ref{fig:qual} shows six images drawn at random with a fixed seed (two per dataset; seed-42 models), not selected by outcome. The three variants produce visually similar restorations, and none is consistently best in per-image PSNR across the six examples. The physics module adds 2.25\,M parameters (\ParamsFull{} vs.\ \ParamsUnet\,M, $+30\,\%$) and raises the cost from \GflopsUnet{} to \GflopsFull\,GFLOPs ($3.4\times$) at $256\times256$. Early stopping ended training after \EpochsFull{} epochs on average for the full model, against \EpochsNoPhys{} (w/o physics), \EpochsNoTask{} (w/o task loss) and \EpochsUnet{} (plain U-Net).

\section{Discussion and limitations}
\label{sec:discussion}

\paragraph{What the evidence supports.} (i) With ample data the physics module is not worth its cost: zero-shot differences are at most 0.16\,dB on LSUI and 0.07\,dB on AUIED3K with intervals that include zero, and the module multiplies the FLOPs by 3.4. (ii) With scarce data the module helps, by up to \DFDiffTenAUIED\,dB on AUIED3K and \DFDiffTenUIEB\,dB on UIEB-val at 80 images, and in the most extreme setting (80 images, AUIED3K) it also lowers the seed-to-seed spread; the trend is consistent but, with five seeds, not individually significant after correction. (iii) Capacity is not the explanation: at 80 images the full model was ahead of a control with the same encoder, decoder and output channels but no formation model by \FGDiffTenAUIED\,dB on AUIED3K and \FGDiffTenLSUI\,dB on LSUI. (iv) The module's outputs are not physical: transmission does not track measured range beyond a random-range control, backscatter has the expected sign in only \STFracBpos\,\% of the images, and the dark-channel prior behaves more physically. (v) The dependence on degradation severity has opposite sign on the two zero-shot sets, which points to a dataset-specific interaction rather than a physical benefit.

\paragraph{Why can an unvalidated physics module still help?} Statements (ii)--(iv) together suggest that the formation-model inversion and its regularizers act as a \emph{structured prior on the computation}: $\hat J_{\mathrm{phys}}$ is a smooth, input-dependent contrast and color stretch that is easy to learn and gives the backbone a strong starting point when little data are available, and the bounded heads and smoothness terms constrain the module's outputs. This is a hypothesis, not a result: our control removes the inversion and the regularizers jointly, and we did not vary the weight of $\mathcal{L}_{\mathrm{phys}}$. What the data do say is that the benefit does not require, and does not come with, physically valid maps.

\paragraph{Scope of the physical validation.} Sea-thru images are linear and were exposure-normalized and sRGB-encoded for the networks, which were trained on display-referred UIEB images; the domain shift may depress any correlation with range, and the module was never supervised with range. Our negative result therefore concerns this \emph{unsupervised} design under this transfer. A module trained with depth supervision, or a physics-based synthesis of training data, may well behave differently; our protocol (correlation with measured range, null control, classical baseline) can be applied unchanged to such designs.

\paragraph{Limitations.}
\begin{itemize}\setlength\itemsep{1pt}
\item One architecture, one training set of 800 images and one resolution ($256\times256$); the conclusions concern this design and regime, not physics guidance in general.
\item Five seeds: seed-level intervals have half-widths of roughly 0.2\,dB on LSUI and 0.5--0.7\,dB on AUIED3K at full data and are wider at low data, so smaller effects cannot be excluded. Holm correction over the 12 data-efficiency comparisons leaves only one significant.
\item Checkpoints are selected on UIEB-val, which is therefore optimistic; zero-shot sets are unaffected. The data-efficiency study uses fixed random subsets of 1000 images of LSUI and AUIED3K.
\item Only full-reference PSNR and SSIM are reported for the zero-shot sets and no downstream task was evaluated, so a benefit that those would reveal is not excluded. LSUI references are high-fidelity references prepared by the dataset authors rather than measurements, and AUIED3K inputs are synthetic degradations of real photographs.
\item The Sea-thru analysis uses at most 100 images per site and evenly spaced frames, which are strongly correlated within a site; confidence intervals are over images and may be optimistic.
\item The severity analysis is post hoc and exploratory, and input PSNR is only a proxy for physical degradation.
\item Loss weights were not re-tuned per variant, and no external state-of-the-art method was compared; we make no claim on absolute competitiveness.
\end{itemize}

\paragraph{Recommendations.} For studies of physics-guided restoration we suggest (a) reporting several seeds with paired intervals instead of one run; (b) evaluating zero-shot on complete external sets rather than on the selection split; (c) including a capacity-matched control without the physical structure; (d) validating predicted physical maps against measured range, with a null control and a classical baseline, before calling them physical; (e) reporting data-efficiency curves, where a prior is most likely to matter; and (f) verifying gradient flow through every loss term before ablating it.

\section{Reproduction note and implementation audit}
\label{sec:repro}
\paragraph{Re-running the published configuration.} Re-training the published single-seed configuration (seed 42, identical split, code after the corrections below) on different hardware (RTX~5080 instead of RTX~5060) and with persistent data-loader workers gave 23.31\,dB PSNR on UIEB-val and 21.53\,dB on the 400-image LSUI subset used in \cite{fauzan2026}, against the reported 22.97\,dB and 21.20\,dB. The run stopped by early stopping at epoch 64 (best epoch 49). A difference of about $+0.3$\,dB between two nominally identical runs is of the same size as most of the effects examined in Section~\ref{sec:results}, which is why we do not interpret single-run ablations.

\paragraph{Defects that invalidate ablations.} A code-level audit of the earlier implementation found five issues that must be excluded before a loss ablation means anything; all are corrected in the version used here and in \cite{fauzan2026}:
\begin{enumerate}\setlength\itemsep{1pt}
\item the frozen task encoder and the VGG feature extractor were wrapped in \texttt{torch.no\_grad()}, so $\mathcal{L}_{\mathrm{task}}$ and $\mathcal{L}_{\mathrm{perc}}$ produced no gradient with respect to the restored image and could not influence training;
\item the custom SSIM implementation could return values above~1, giving negative loss terms;
\item the frozen task encoder was executed at inference although it is training-only;
\item ablation variants were not re-seeded before model construction, so their initial weights differed;
\item the evaluation script instantiated learned baselines without loading their trained weights.
\end{enumerate}
A sanity check now asserts that the restored image requires gradients and that every loss term back-propagates to it before any training run is started. We recommend such a check for any study that ablates loss terms computed on frozen feature extractors.

\section{Conclusion}
We asked whether the physics in a physics-guided underwater restoration network does any work, and whether it is physics. Under controlled conditions the answer is nuanced. With ample data the physics module brings no zero-shot accuracy benefit and costs $3.4\times$ the FLOPs. With scarce data it helps, and a capacity-matched control without the formation model does not reproduce this, so the benefit is tied to the physics-shaped structure (a consistent trend, though not individually significant with five seeds). Yet the predicted transmission does not follow measured range, and the effect of the module changes sign with degradation severity between datasets. The physics here is an inductive bias, not a measurement. The practical consequence is twofold: a physical prior is worth its cost when data are scarce, and the maps it produces should not be described, or used, as physical quantities without validation against range. More generally, single-run ablations on small benchmarks can reverse in sign; multi-seed, zero-shot reporting, capacity-matched controls and range validation are inexpensive safeguards that we hope become routine.

\paragraph{Data and code availability.} UIEB, LSUI, AUIED3K and Sea-thru are public datasets \cite{li2020uieb,peng2023,auied3k,akkaynak2019}. Code, configurations and per-image results are available from the corresponding author on request.

\end{document}